\documentclass[10pt|11pt|12pt]{article}
\usepackage{cite}
\usepackage{amssymb,amsmath,latexsym,mathrsfs}
\usepackage{psfrag}
\usepackage{bm}
\usepackage{cite}
\usepackage{url}
\usepackage{color}
\usepackage{graphicx}
\usepackage{caption}
\usepackage{subcaption}
\usepackage{booktabs}
\usepackage{multirow} 
\usepackage{mathtools} 
\usepackage{extarrows} 
\usepackage{centernot}

\usepackage{authblk}

\usepackage{cite}
\usepackage{amsthm}
\newtheorem{defi}{Definition}
\newtheorem{theorem}{Theorem}[section]
\newtheorem{thm}{Theorem}[section]
\newtheorem{lemma}[theorem]{Lemma}

\newtheorem{corollary}[theorem]{Corollary}

\newtheorem{example}{Example}

\newtheorem{compsol*}{Complete Solution}

\usepackage{datetime}

\newdate{date}{26}{02}{2020}
\date{\displaydate{date}}

\begin{document}

\title{\LARGE \bf Decidability of Sample Complexity of PAC Learning in finite setting}

\author{ Alberto Gandolfi }

\affil{NYU Abu Dhabi}

\vspace{5mm}

% I decided to use the words   "ASSUMPTIONS"   and "CONDITIONS"
%
%
%

\maketitle

\thispagestyle{plain}
\pagestyle{plain}


\begin{abstract}
In this short note we observe that the sample complexity of PAC machine learning of various concepts, 
including learning the maximum (EMX), can be exactly determined when
the  support of the probability measures considered as models satisfies an a-priori bound.
This result contrasts with the 
 recently discovered undecidability
 of EMX  within ZFC
 for finitely supported probabilities (with no a priori bound).
Unfortunately, the decision procedure is at present, at least doubly
exponential in the number of points times the uniform bound on the
support size.

\end{abstract}

\section{Introduction} \label{1}

%New Version

It has been recently discovered
\cite{Ben-David2019}
that machine learnability can be undecidable within ZFC  (Zermelo-Fraenkel with axiom of Choice). For a given degree of 
approximation $\epsilon$ and value of residual probability
$\delta$, learnability of 
a concept over a certain class $\mathcal P$ of probabilities
consists of the existence of a learning algorithm, a learner, and a number $m$
such that with $m$ independent observations from anyone of the possible 
probabilities $P \in \mathcal P$, the concept is approximated by the 
learner within an error of at most $\epsilon$ with 
$P$ probability larger than $1-\delta$.
\cite{Ben-David2019} shows that there are situations in which
learnability of a concept is independent of ZFC 
axioms systems, even restricting to finitely supported
probabilities. 
It thus becomes  relevant to explore conditions which ensure decidability of  learning.

%To explore decidability, it is convenient to look at things from
%the opposite side: given a sample size $m$, 
%an algorithm is not an $(\epsilon, \delta)$-learner if there exists a probability $P $, in the class $\mathcal P$, which violates, with $m$ independent observations, the degree of approximation with a probability
%exceeding the
%bound on residual probabilities (see Section $5$ (??) below). The problem has thus been turned
%into that of existence of a probability satisfying certain additional
%conditions, or assumptions, besides those required for it
%to be a probability.
%
%The requirement that some relation holds for all probabilities in a certain class 
%corresponds exactly to the requirement that a formula is valid in the semantic interpretation of 
%a logic: semantic validity consists exactly of the fact that the formula is true for all 
%structures. The logic we need here is one which allows to express 
%the assertion that a concept is approximated by the 
%learner within an error of at most $\epsilon$ with 
%$P$ probability larger than $1-\delta$, and whose semantic structures
%consist of a certain class of probabilities.

To this extent, it is convenient to look at things from the 
 opposite side: given a sample size $m$, 
an algorithm is not an $(\epsilon, \delta)$-learner if there exists a probability $P $, in the class $\mathcal P$, which violates, with $m$ independent observations, the degree of approximation with a probability
exceeding the
bound on residual probabilities. This way the problem has  been turned
into that of existence of a probability, within a certain class,  satisfying certain additional
conditions, besides those required for it
to be a probability (and to be in the prescribed class).

To explore decidability, we  consider   the so called "discretization trick" (\cite{S-SB-D2014} - Remark 4.1), 
according to which in virtually all concrete applications various limitations,
such as using a computer to handle the data,
introduce an a-priori bound on the number of possible states
of the system. 
We then show that in this case, learnability can be expressed in terms of 
polynomial relations for the probabilities of specific events;  we can then refer
to the Tarski-Seidenberg Theorem for real closed fields \cite{BCR}, which 
shows that existence of a solution for any finite set of polynomial relations
is decidable. This provides an explicit, albeit computationally very expensive,
algorithm for the determination of the sample complexity.
Very accurate bounds have been developed for the sample complexity \cite{H16},
but, in view of \cite{Ben-David2019}, they provide no guarantee of decidability.

It is interesting to realize that the Tarski-Seidenberg Theorem does not indicate
how to find a probability violating a given tentative learner $G$, nor could it provide 
any exact method for it, as in general there is no finite algorithm to determine solutions
of polynomial equations of degree greater than or equal to $5$. 
It is also interesting to notice that the theories of natural  \cite{G} or rational numbers \cite{R49}
are
not decidable; so, if we insisted in restricting to probabilities taking rational values it would
not be clear if learnability is decidable. 
On the other hand,
we are not interested in any such restriction, or in the exact determination of the probabilities, but only in 
their
existence, in order to exclude a tentative learner $G$, or non existence,
in order to assess that $G$ is a learner; and that's exactly what the
Tarski-Seidenberg Theorem guarantees. In other words, decidability is guaranteed
if we discretize the inputs, but not the values of the modeling probabilities.

As a related topic, we mention that the issue of existence of a probability satisfying certain requirements
can be given an interpretation in terms of Logic. In this context, one
develops first the syntax of a logic, i.e. the allowed symbols and formulas;
then, the collection
of models in which the formulas can  be interpreted represents  the
semantics  of a logic. When models are probabilities of a certain class, then
a formula is valid if it is true for all probabilities of the class: a logic for finite
probabilities with rational coefficients has been developed in \cite{FHM1990}. Learnability can then
be interpreted in terms of validity of the formulas expressing
the fact that a certain $G$ is a learner. We briefly discuss and exploit this connection further in Section
\ref{Complexity}.

%Notice that Tarski Seidenberg result within the theory of real closed fields
%is interpretable in ZFC

%Although interesting from the abstract point of view, especially in relation 
%to the the lack of decidability observed by \cite{Ben-David2019}, 
%the 

%
%
%Once   the support of all the probabilities in the class is identified as a finite set, 
%there is a logic developed, among other related logics, in [Fagin, Halpern, Megiddo 1990],
%which includes the formulas for learnability, and  has probabilities with, effectively, 
%a given finite support as structures.
%It is shown there that
%validity is decidable by Tarski Seidenberg Theorem. In order to simplify
%and streamline the exposition, 
%we recap here the 
%results about decidability, without developing the details of the syntax of
%the logic.
%
%We then proceed with a simple algebraization of  PAC-learnability which allows to exploit the Tarski-Seidenberg reduction
%theorem, to show exact decidability. Exact decidability consists in the exact determination
%of the learning dimension, as opposed to the usual upper bound.

\section{Polynomiality of learnability conditions and sample complexity} \label{declearn}

Sample complexity of agnostic PAC learning
 is defined as follows. Let 
$\mathcal X, \mathcal Y$ be  sets, indicating the set of features and labels, respectively, and let $\mathcal H=\{h:\mathcal X \to \mathcal Y\}$
indicate a hypothesis set.
To evaluate a hypothesis, we introduce a loss
$
L: \mathcal Y\times \mathcal Y \to \mathbb R^+$, and a loss function
$\ell: \mathcal H\times \mathcal X \times \mathcal Y$ defined by
$\ell(h, (x,y)):=L(h(x), y)$. When a probability $P$
is defined on $\mathcal X\times \mathcal Y$,
and $X, Y$ are two random variables taking values in 
$\mathcal X$ and $\mathcal Y$, respectively, with joint 
distribution $P$, the average loss of
a hypothesis $h$ is $\mathcal E_P(h):=
E_{P}(\ell(h,(X,Y))$. For $m \in \mathbb N$,
a potential $m$-learner is a function $G:(\mathcal X\times \mathcal Y)^m \to \mathcal H$, where  we denote
$h_{G, (x_1, y_1), \dots, (x_m, y_m)}:=G(((x_1, y_1), \dots, (x_m,y_m)))$.
Given a class $\mathcal P^*$ of probabilities on  (possibly suitable subsets of) $\mathcal X
\times \mathcal Y$, and  $ \epsilon, \delta >0$, an $(m,\epsilon,\delta)$-learner  with respect to $\mathcal P^*$ is a
potential $m$-learner $G$ such that 
\begin{equation}\label{10.1}
 P^n(E_P(\ell(h_{G,(x_1, y_1), \dots, (x_m, y_m)}, (X,Y)))\leq \min_{h' \in \mathcal H}
\mathcal E_P(h') + \epsilon ) \geq 1-\delta
\end{equation}
for all $P \in \mathcal P^*$.
The class $\mathcal H$ is {\bf agnostically PAC learnable} with respect to $\mathcal P^*$
if for every $0<\epsilon, \delta  \in \mathbb R $ there exists $m(\epsilon, \delta)$ such that for all $m \geq m(\epsilon, \delta) $
there exists an $(m,\epsilon,\delta)$-learner $G$. The 
{\bf sample complexity} of 
 agnostically PAC learning the
class $\mathcal H$  with respect to $\mathcal P^*$
is the minimum $m_{\mathcal H}$ of such
$m(\epsilon, \delta) $'s.
%%%%%% The theorem %%%%
\begin{thm} \label{15.1}
The sample complexity of  agnostic PAC learnability with respect to the class $\mathcal P^*$ of all probabilities
with  uniformly bounded support  is decidable.
\end{thm} 
\begin{proof} 
Suppose the support of each probability is bounded by some uniform constant $\overline n$,
and consider first fixed finite sets 
 $\mathcal X$ and $\mathcal Y$ such that
 $|\mathcal X \times \mathcal Y| \geq  \overline n$. In this case, 
the hypothesis class $\mathcal H=\{h'| h':\mathcal X \to \mathcal Y\}$ is finite as well. A finite hypothesis 
class is learnable (\cite{S-SB-D2014} Cor 4.6), hence
 for $ \epsilon, \delta >0$ there exists $m(\epsilon, \delta)$ such that for all $m \geq m(\epsilon, \delta) $
  \eqref{10.1} holds for some $G$.
  
  Next, for each $m< m(\epsilon, \delta)$, potential
$m$-learner $G$, and hypothesis $h' \in \mathcal H$, we say that
$G$ is not a $(m,\epsilon,\delta)$-learner of $h'$ if 
there exists a probability $P$ on $\mathcal X \times \mathcal Y$ such that 
\begin{equation}\label{10.2}
 P^n(E_P(\ell(h_{G, (x_1, y_1), \dots, (x_m, y_m)}, (X,Y)))> 
\mathcal E_P(h') + \epsilon ) > \delta;
\end{equation}
 notice that $G$ is not a $(m,\epsilon,\delta)$-learner if there is an $h'\in \mathcal H$
such that $G$ is not a learner of $h'$; we also say that a probability $P$ satisfying \eqref{10.2} violates $G$.
Condition \eqref{10.2} is polynomial in the probabilities $P(A)$'s of some events $A \subseteq \mathcal (X \times \mathcal Y)^m$, 
in the following sense: for $j=1, \dots, m$, let $ (\mathcal X \times \mathcal Y)^{(j)}$
be  a copy of $ \mathcal X \times \mathcal Y$, and
consider events $A_{(x,y)}^{(j)}=\{X_j=x, Y_j=y\}, (x,y)\in(\mathcal X \times \mathcal Y)^{(j)}$, 
 where $X_j, Y_j$ are the realizations of the $j$-th trial;
next, let $p^{(j)}_{(x,y)}:=P(A^{(j)}_{(x,y)})$. Then, given $h'$, the existence of a probability $P$ satisfying \eqref{10.2}
is easily seen to be equivalent to the existence of a solution of
\begin{eqnarray} \label{10.21} 
\begin{cases}
\sum_{ (x_1, y_1), \dots, (x_m, y_m) \in \mathcal X\times \mathcal Y:
\mathcal E_P(h_{G, (x_1, y_1), \dots, (x_m, y_m)}) >\mathcal E_P(h') + \epsilon} 
p_{(x_1, y_1), \dots, (x_m, y_m)} > \delta \\
p_{(x_1, y_1), \dots, (x_m, y_m)} = \prod_{j=1}^m
p^{(j)}_{(x_j,y_j)}\\
p^{(j)}_{(x,y)}=p^{(1)}_{(x,y)}, \text{ for } (x,y) \in (\mathcal X \times \mathcal Y)^{(j)},  j=2, \dots, m\\
\sum_{(x,y) \in \mathcal X \times \mathcal Y}
p^{(j)}_{(x,y)}=1, \text{ for }j=1, \dots, m \\
p^{(j)}_{(x,y)} \geq 0, \text{ for } (x,y) \in (\mathcal X \times \mathcal Y)^{(j)}, j=1, \dots, m.
\end{cases}
\end{eqnarray}
System \eqref{10.21} is written in terms of unknowns $p^{(j)}_{(x,y)}$'s using polynomials
and indicator functions. In order to apply the results of the next section, we need to 
eliminate the indicator functions. This can be done by considering subsets $A
\subseteq  (\mathcal X \times \mathcal Y)^m$, and 
then observing that there is a solution
to \eqref{10.21}  if and only if there is a solution to at least one  of the systems
in the following collection
labeled by $A:
\subseteq  (\mathcal X \times \mathcal Y)^m$,
\begin{eqnarray} \label{10.3} 
\begin{cases}
\sum_{ (x_1, y_1), \dots, (x_m, y_m) \in A}
p_{(x_1, y_1), \dots, (x_m, y_m)} > \delta \\
\mathcal E_P(h_{G, (x_1, y_1), \dots, (x_m, y_m)}) >\mathcal E_P(h') + \epsilon,
 \text{ for }  ( (x_1, y_1), \dots, (x_m, y_m)) \in A\\
 \mathcal E_P(h_{G, (x_1, y_1), \dots, (x_m, y_m)}) \leq \mathcal E_P(h') + \epsilon,
 \text{ for }  ( (x_1, y_1), \dots, (x_m, y_m)) \notin A\\
p_{(x_1, y_1), \dots, (x_m, y_m)} = \prod_{j=1}^m
p^{(j)}_{(x_j,y_j)}\\
p^{(j)}_{(x,y)}=p^{(1)}_{(x,y)}, \text{ for } (x,y) \in (\mathcal X \times \mathcal Y)^{(j)},  j=2, \dots, m\\
\sum_{(x,y) \in \mathcal X \times \mathcal Y}
p^{(j)}_{(x,y)}=1, \text{ for }j=1, \dots, m \\
p^{(j)}_{(x,y)} \geq 0, \text{ for } (x,y) \in (\mathcal X \times \mathcal Y)^{(j)}, j=1, \dots, m.
\end{cases}
\end{eqnarray}
 Since $\mathcal E_P$ is an expected value,
hence a linear condition, all the relations in each of the systems of the form \eqref{10.3} are polynomial
in the variables $p^{(j)}_{(x,y)}$'s.
We then have at most $2^{|\mathcal X \times \mathcal Y|^m}(|\mathcal X \times \mathcal Y|^ {m(\epsilon, \delta)}+m(\epsilon, \delta)
|\mathcal X \times \mathcal Y|+m+1)$ 
polynomial conditions to check. For each $m< m(\epsilon, \delta)$, this is  is decidable within the theory of real-closed fields
by  Theorem \ref{Decidability} below. Hence 
the sample complexity of agnostic PAC learnability, with respect to set $\mathcal P^*$ of all probabilities on $\mathcal X
\times \mathcal Y$,
is  decidable.

Finally, consider any set $\overline  {\mathcal X} $,  a finite set of labels 
$\tilde {\mathcal Y}$, and one of the finite subset $\tilde {\mathcal X} \subseteq \overline  {\mathcal X} $
such that  $|\tilde {\mathcal X} \times \tilde {\mathcal Y}| \leq \overline n$;
 $\tilde {\mathcal X} \times \tilde {\mathcal Y}$ is viewed as a possible support of a
probability violating a potential $m$-learner $G$. Consider also an injective maps from
$\tilde {\mathcal X}$ to $\mathcal X$, and from $ \tilde {\mathcal Y}$ to $  \mathcal Y$,
respectively, where $\mathcal X$ and $\mathcal Y$ are the fixed sets considered above.
The action of $G$ on observations from 
$\tilde {\mathcal X} \times \tilde {\mathcal Y}$ and the existence of a probability $P$ on 
$\tilde {\mathcal X} \times \tilde {\mathcal Y}$
violating $G$ is determined by systems of the form \eqref{10.3},
which are preserved by the above injective maps (assigning probability zero to all points in 
$\mathcal X \times \mathcal Y$ not in the image of $\tilde {\mathcal X} \times \tilde {\mathcal Y}$).
So, whether $G$ is a learner or not for given $m, \epsilon, \delta$ can be determined by the
fact that the there is a learner or not on $\mathcal X \times \mathcal Y$,
which we have seen is decidable. Hence, sample complexity  of  agnostic PAC learnability with respect to the class $\mathcal P^*$ of all probabilities
with   support  uniformly bounded by $\overline n$  is decidable 

\end{proof}

In general, the exact value of the sample complexity can be only determined by a systematic examination, 
which is  guaranteed to end by the above theorem.
\begin{example}
Consider learning the maximum in a binary space $\mathcal X $,
which we can then assume to be $\mathcal X=\{0,1\} $. Suppose   that 
the hypothesis class is $\mathcal H=\{\{0\},\{1\}\}$ and we use the ERM learner. 

For 
$\epsilon=1/3$ and $ \delta=1/3$ we have $m_{\mathcal H}=1$, while the standard upper bound based on Hoeffding theorem
gives $m_{\mathcal H} \leq \lceil 2\log{(2 |\mathcal H|/\delta)}/\epsilon^2 \rceil=45$.

For 
$\epsilon=1/10, \delta=1/100$, the standard upper bound 
gives $m_{\mathcal H} \leq 1199$;
 some known lower bounds give $m_{\mathcal H} \geq 41$,
and there is a matching upper bound but with an unkonwn constant
\cite{H16}; the sample complexity
  turns out to be $m_{\mathcal H}
=539$. 
\end{example}

%%%%%%

\section{Decidability of finitely many
polynomial problems in finite probabilities
 by Tarski-Seidenberg Theorem}
 
 System \eqref{10.3} is a special case of a general situation which occurs often in 
elementary probability: there are
an unknown probability
$P$;
 a finite number of events, $A_1, \dots, A_n$;  and then a finite number $R$ of polynomial relations 
 that are to be satisfied by
  the probabilities of 
either the $A_i$'s or some of their boolean combinations. 
 
 Possibly using the  disjunctive normal form, one can always reduce these problems to
 a collection of polynomial relations, equalities and inequalities, in variables
 which represent the probabilities of a finite set $S$ (or, equivalently, in terms of
 the probabilities of the atoms of the normal form).
 
 Expressed in general terms, we  arrive at
a system of polynomial relations in the variables
$p_{ \alpha}, \alpha \in S=\{s_1, \dots , s_{|S|}\} $
of the form
\begin{eqnarray} \label{3}
\begin{cases}
g_r( {\bf p}) \triangleleft 0, \quad r=1, \dots, k-S-1 \\
p_{ \alpha(s)} \geq 0, \quad s=1, \dots, S \\
\sum_{ \alpha \in S}
p_{ \alpha}=1
\end{cases}
\end{eqnarray}
for some $k \in \mathbb N$, where $g_r$'s are polynomials, ${\bf p}=
(p_{s_1}, \dots, p_{s_{|S|}})$, and $\triangleleft$ stands for either of $\geq, =, \neq$.

Algebraic Geometry has developed the appropriate tools to 
decide if such a system has a solution (see, e.g.. \cite{BCR} or \cite{BPR}).
\begin{thm} \label{Decidability}
The existence and nonexistence of probabilities
satisfying system \eqref{3},  is decidable.
 
\end{thm}
\begin{proof}
System \eqref{3}
determines a semi-algebraic set, which is nonempty if and only
if there are solutions satisfying all the equations. Whether a semi-algebraic set
is empty or not is decidable with the following
 decision procedure. First, by the Tarski-Seidenberg Theorem
 a semi-algebraic 
set in $\mathbb R^{k+1}, k\geq 2,$ is non empty if and only if
its projection on $\mathbb R^{k}$ is non empty
(see e.g. \cite{BPR}, Theorem $2.76$); iterating, this procedure
reduces the problem to semi-algebraic sets in $\mathbb R$.
For these, every semi-algebraic
set can be decomposed in finitely many basic semi-algebraic sets
of the form $\{x \in \mathbb R| P(x)=0, Q(x)>0 \text{ for all }
Q \in \mathcal Q\}$, where $P$ is a polynomial, and 
$\mathcal Q$ is a collection of polynomials. 
Finally, whether each basic semi-algebraic set is non empty
can be determined  by a General Law of Signs, 
which consists of checking the signs of suitable combinations of the 
coefficients of the polynomials (see e.g. \cite{BPR}, Lemma $2.74$)
\end{proof}.
The last step is similar  to the methods in Sturm's Theorem
or Descartes€ Law of Signs.

\section{Computational complexity} \label{Complexity}

It is shown in \cite{FHM1990}, Theorem $5.3$ that when the coefficient of the 
polynomials are rational, as they would be in any implementation,
there is a procedure, for deciding if a polynomial weight formula
is satisfiable in a (finite) probability space, that runs in polynomial space.
It is then easy to see that 
 each of the systems
\eqref{10.3} can be expressed as a polynomial weight formula
in the language of  \cite{FHM1990}, Chapter $5$: as set of primitive
propositions we take $\Phi= (\mathcal X \times \mathcal Y)^m$
with boolean operations defined as usual for subsets of 
$(\mathcal X \times \mathcal Y)^m$; weight terms 
are $w( ((x_1, y_1), \dots, (x_m, y_m)) ):= p_{(x_1, y_1), \dots, (x_m, y_m)} $,
with linear operations and multiplications allowed to make formulas.
The semantics to these weight formulas is then given by 
probabilities on the set $(\mathcal X \times \mathcal Y)^m$,
and hence \cite{FHM1990}, Theorem $5.3$ applies
to the decision problem of each of the systems  \eqref{10.3}.

In terms of number of arithmetic operations, on the other hand,
the implementation of Tarski-Seidenberg elimination and 
the General Law of Signs has  very high complexity; a slightly better
version is cylindrical decomposition (see e.g. \cite{BPR},
Ch. $5$), which is implemented in various software, but
remains doubly exponential in the number of variables and 
of equations: for the System \eqref{3} it takes $(k m)^{2^{\mathcal O( S)}}$    \cite{B17}
operations to decide whether a solution exists. So, the direct calculation of the
sample complexity
 using this method is accessible
only for problems with a very small a-priori bound $\overline n$..

Expressed in terms of the number of pixels and colors in an image
the number of arithmetic operations 
needed to determine the sample complexity of learning a hypothesis class
$\mathcal H$
would be a quadruple exponential; 
something of the order of $|\mathcal H| 2^{2^{10^{5m}}}$ for a $64\times 64$, $16$-color image and 
a sample of size $m$.

\section{Discussions and conclusions}

We  make noe a partial exploration of  the source of the undecidability
found in \cite{Ben-David2019}
 when learning is seen from the point of view of 
existence/nonexistence of probabilities satisfying suitable conditions. This examination is hindered in 
\cite{Ben-David2019}, as in that paper learning is equivalently expressed in terms of compression schemes.
The onset of undecidability is partially elucidated by the following.

\begin{corollary}
Let $G$ be a $(\overline m,1/3,1/3)$-learner  of EMX for finitely supported probabilities defined
on a model $M$ of ZFC satisfying CH; when extended to a 
model $M'$ containing $M$ and satisfying $\neg CH$,  $G$  determines a system of the form \eqref{10.3},
with $m=\overline m$,
which admits a solution. 

\end{corollary}
One example of such extension is
obtained from the use of the forcing method \cite{C63}.
 
\begin{proof}
Consider a model $M$ of ZFC satisfying CH.
It is shown in \cite{Ben-David2019} that  there is an $\overline m$ and  a $(\overline m, 1/3, 1/3)$
learner $G$ of EMX over the collection of finitely supported probabilities in $[0,1]$. For any given finite collection  $s_1, \dots, s_n \in [0,1]$ which could be used as support of 
a finite probability violating $G$, $G$ determines  a finite number of systems of the form
\eqref{10.3}, with $m=\overline m$, none of which has 
a solution (since $G$ is a learner).

Consider now an extension $M'$ of $M$ satisfying $\neg CH$.
Suppose the learner $G$ is extended to the sequences
$( (x_1, y_1), \dots, (x_n, y_n))$ such that some of the $x_i$'s  do not belong to M.  Then the existence of a probability violating
the extension of $G$ is also determined by systems of the form 
\eqref{10.3}, with $m=\overline m$, but now there must be a solution
for at least one of such systems, as the extension of $G$ cannot
be a learner.

\end{proof}

\bigskip
To summarize the results of the paper, we have shown that,
aside from the very high computational complexity of the decision procedure,
the exact determination of the sample complexity in agnostic PAC learning,
including EMX, is decidable under  the "discretization trick" (i.e. when the probabilities are known to be supported on a finite set with an a-priori bounded size). This 
result contrasts with the undecidability of learning discovered in \cite{Ben-David2019} for learning the maximum
with $\epsilon=\delta=1/3$ with respect to probabilities supported on a finite set (whose size has no a priori bound).
We have also investigated the mechanism by which a learner developed 
in a model satisfying CH fails in any extension to a model in which CH ceases to hold.

 %*****************************************

Contact address:
NYU Abu Dhabi 
Saadiyat Island
P.O Box 129188
Abu Dhabi, UAE

email: ag189@nyu.edu


\small 
\begin{thebibliography}{99}



\bibitem[Ben-David et al. 2019]{Ben-David2019}  Shai Ben-David, Pavel Hrubesss, Shay Moran, Amir Shpilka, and Amir Yehudayoff.  Learnability can be undecidable.
Nature Machine Intelligence {\bf 1}, 1 (2019), 44-48.


\bibitem[Bochnak, Coste, Roy 1998]{BCR}   Bochnak, Jacek; Coste, Michel; Roy, Marie-Franccoise. Real Algebraic Geometry. Translated from the 1987 French original. Revised by the authors. Ergebnisse der Mathematik und ihrer Grenzgebiete (3) [Results in Mathematics and Related Areas (3)], 36. Springer-Verlag, Berlin, 1998.

\bibitem[Basu 2017]{B17} Saugata Basu. 2017. ALGORITHMS IN REAL ALGEBRAIC GEOMETRY:
A SURVEY. Panoramas \& Synthèses, 51, 2017, 107-153.

\bibitem[Basu, Pollack, Roy 2006]{BPR} Saugata Basu, Richard Pollack, and Marie-Franccoise Roy. 2006. Algorithms in Real Algebraic Geometry (Algorithms and Computation in Mathematics). Springer-Verlag New York, Inc., Secaucus, NJ, USA.

\bibitem[Cohen 1963]{C63} Cohen, Paul J., 1963. The Independence of the Continuum Hypothesis". Proceedings of the National Academy of Sciences of the United States of America. 50 (6): 1143-1148.

\bibitem[E2006] {E} Richard L. Epstein Classical Mathematical Logic: The Semantic Foundations of Logic, Princeton University Press.

\bibitem[Fagin, Halpern and Megiddo 1990]{FHM1990}  Fagin, R., Halpern, J. H. and Megiddo N:
A Logic for Reasoning about Probabilities, Inform. and Comput., 87, Nos. 1/2, 1990

\bibitem[G\"odel 1931]{G}  Kurt G\"odel (1931), "\"Uber formal unentscheidbare Sätze der Principia Mathematica und verwandter Systeme, I." Monatshefte f\"ur Mathematik und Physik 38, 173-198.

\bibitem[GKP1988]{GKP1988}G. Georgakopoulos, D. Kavvadias, and C. H. Papadimitriou. Probabilistic satisfiability. Journal of Complexity, 4:1-11, 1988. 

\bibitem[Hanneke 2016]{H16}
S. Hanneke: The Optimal Sample Complexity of PAC Learning.
Journal of Machine Learning Research 17 (2016) 1-15.


\bibitem[HM2001]{HM} Hazewinkel, Michiel, ed. (2001), "Disjunctive normal form", Encyclopedia of Mathematics, Springer,

\bibitem[Robinson 1949] {R49} J. Robinson (1949). Definability and decision problems in 
arithmetic. The Journal of Symbolic Logic, 14, 98-114.


  \bibitem[Shalev-Shwartz and Ben-David 2014]{S-SB-D2014} Shalev-Shwartz, S. and Ben-David, S. (2014) Understanding Machine Learning: From theory to algorithms. Cambridge: Cambridge University Press.  %
 
\end{thebibliography}
\end{document}